\documentclass[conference]{IEEEtran}
\IEEEoverridecommandlockouts

\usepackage{cite}
\usepackage{amsmath,amssymb,amsfonts}
\usepackage{algorithmic}
\usepackage{graphicx}
\usepackage{textcomp}
\usepackage{xcolor}
\usepackage{float}
\usepackage{tikz}
\usepackage{placeins}

\begin{document}

\title{Marine Snow Removal Using Internally Generated Pseudo Ground Truth

\thanks{This work is supported by EPSRC IAA, University of Bristol, and EPSRC ECR (EP/Y002490/1).}}




\author{Alexandra Malyugina$^1$, Guoxi Huang$^1$, Eduardo Ruiz$^2$, Benjamin Leslie$^2$ and Nantheera Anantrasirichai$^1$ \\

\IEEEauthorblockA{$^1$\textit{Visual Information Laboratory}, \textit{University of Bristol}, Bristol, UK \\
$^2$\textit{Beam}, Bristol, UK 
}}


\maketitle

\begin{abstract}
    Underwater videos often suffer from degraded quality due to light absorption, scattering, and various noise sources. Among these, marine snow—suspended organic particles appearing as bright spots or noise—significantly impacts machine vision tasks, particularly those involving feature matching. Existing methods for removing marine snow are ineffective due to the lack of paired training data. To address this challenge, this paper proposes a novel enhancement framework that introduces a new approach to generating paired datasets from raw underwater videos. The resulting dataset consists of paired images of generated snowy and snow-free underwater videos, enabling supervised training for video enhancement. We describe the dataset creation process, key characteristics, and demonstrate its effectiveness in enhancing underwater image restoration with the absence of groundtruth.
\end{abstract}


\section{Introduction}
\label{sec:introduction}
Video enhancement plays an important role in many computer vision applications, improving the visibility of objects and features under challenging conditions. It is also an important preprocessing step for tasks such as object detection, tracking, and simultaneous localization and mapping (SLAM), where degraded visual quality can significantly affect algorithm performance. In underwater environments, video degradation may be caused by light absorption and scattering, as well as various types of noise and marine snow. This makes video enhancement a critical preprocessing step for marine robotics, and autonomous vehicles.

Among underwater visibility issues, marine snow is a common and severe problem that can affect the performance of machine vision tasks such as SLAM. Marine snow consists of particles of near-neutral buoyancy suspended in the water column. When collecting video data with artificial illumination marine snow appears as bright spots or noise in the video, leading to quality degradation and obstruction of key features.. Marine snow is highly variable and difficult to model, making its removal challenging. Traditional methods for mitigating marine snow include histogram manipulation and filtering techniques, model-based methods and machine learning-based approaches. However, these approaches have limitations: filtering methods may remove valuable details, histogram methods are hard to tune for optimal performance and deep learning based approaches, despite promising results, often rely on synthetic data that fails to capture the complexity of real-world distortions.

Comprehensive training data is a key component of deep learning based methods. However,  the creation of high-quality paired datasets often requires labor-intensive processes such as sophisticated modelling, data simulation and extensive data cleaning. This problem is particularly challenging  in niche domains like underwater image and video restoration, where data acquisition conditions are inherently complex, and ground truth data is not available.

To address these limitations, we propose a simple dataset creation method that transforms raw, unpaired data captured at various depths into a paired dataset without the need for ground truth. Our approach is based on straightforward but efficient frame blending to simulate paired data directly from raw underwater footage with minimal manual annotation effort. Moreover,  we propose an underwater video enhancement method designed to improve SLAM performance in low-light environments affected by marine snow. Unlike previous approaches that discard or filter the keypoints, our method enhances entire video frames, preserving feature-rich regions while suppressing noise. 

The key contributions of the paper are:
\begin{itemize}
 \item A video enhancement method for marine snow mitigation, improving SLAM performance without explicit keypoint filtering.
\item A data extraction approach that eliminates the need for paired or manually cleaned data, using raw underwater video captured at different depths for training.
\item A demonstration of the method’s effectiveness, showing that enhancing entire video frames improves feature tracking and 3D reconstruction.

    
\end{itemize}

\begin{figure*}
    \centering
    \includegraphics[width=0.95\textwidth]{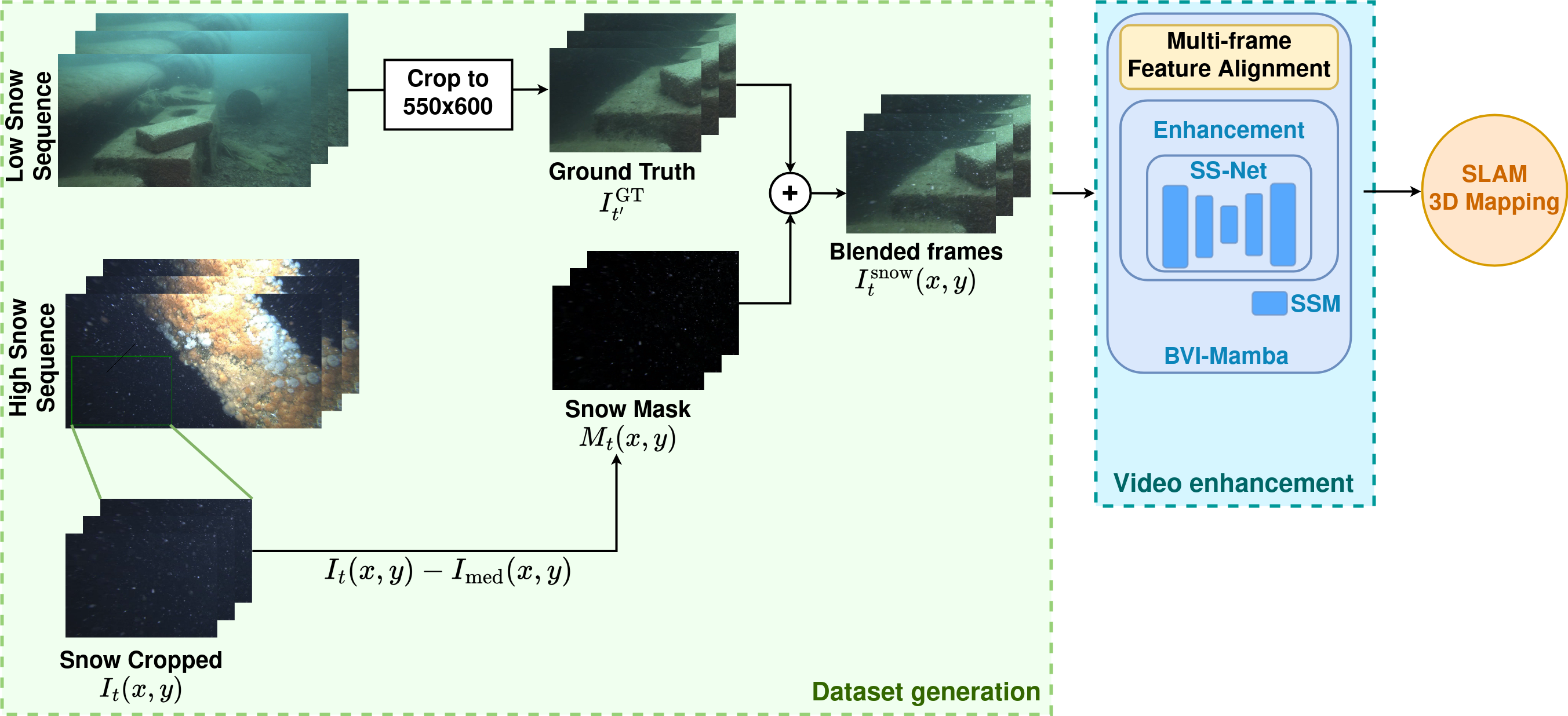}
    \caption{Diagram of the proposed framework}
    \label{fig:diagram}
\end{figure*}

\section{Related Work}
\label{sec:related_work}

Numerous methods have been developed to address distortions caused by atmospheric and underwater particles in videos. Early approaches could predominantly be applied to a single image and relied on handcrafted features such as guidance images, histogram of gradients \cite{xu2012improved,yu2014content,pei2014removing,bossu2011rain} and filtering techniques \cite{ding2016single,xu2012removing}, to separate particles from the background. However these methods come at a cost of oversmoothing and loosing details or do not guarantee the removal of the particles and struggle with temporal consistency when applied to a video sequence. More recent atmospheric snow removal methods are based on Gaussian Mixture Models \cite{bin2020videosnow}, Maximum a Posteriori framework \cite{minghan2021online}. There are also some recent deep learning approaches, such as contrastive learning models with temporal aggregation \cite{chen2023snow}, which, like many others, rely on synthetic data \cite{chen2023snow,coffelt2023marine}.

Despite atmospheric snow removal having been extensively studied for both images and videos, these methods do not fully address the challenges of marine snow removal. Video atmospheric snow removal methods assume that snowflakes, being ruled by gravitation, move in relatively uniform downward trajectories. Underwater snow is composed of suspended organic and inorganic particles and exhibits complex and non-uniform motion influenced by water currents and turbulence. These particles are often large and dense and do not follow simple gravitational paths but move in multiple directions, making their removal considerably more challenging.

Numerous methods have been proposed for image-based underwater enhancement \cite{wang2021underwater,chongyi2020enhabcementdataset,galetto2025uwimggandatasetunet, yiping2024uwrestore, huang2025bayesian}, with some methods focusing on supervised learning trained on synthetic data generated by manual pairwise voting on enhanced images \cite{chongyi2020enhabcementdataset} or a generative network (GAN) \cite{galetto2025uwimggandatasetunet}.  However, research on marine snow removal in underwater videos remains limited. 

Cyganek and Gongola \cite{cyganek2018real} developed a real-time filtering method based on spatiotemporal patch analysis and 3D median filtering, which is effective for remotely operated vehicles (ROVs) but relies on handcrafted rules, limiting adaptability to diverse marine snow conditions. More recently, Coffelt et al.\cite{coffelt2023marine} proposed a fully synthetic dataset for marine snow simulation and trained a UNet-based deep learning model for snow removal. While this synthetic data provides controlled training conditions, the lack of real-world marine snow samples may result in poor generalization. Hodne et. al in \cite{hodne2022detecting} focus on filtering keypoints for SLAM by classifying marine snow in keypoint detection, which requires integrating keypoint rejection into SLAM pipelines and training classifiers. The data generation process involves manual collection of marine snow samples from specific conditions and then superimposing them onto background.

Unlike these methods, our approach uses real underwater video sequences to extract marine snow masks via median frame subtraction, enabling supervised training on realistic data while maintaining temporal consistency.

\section{Methodology}
\label{sec:method}

The overview of our framework is shown in Fig. \ref{fig:diagram}, consisting of dataset generation and video enhancement modules.

\subsection{Dataset Generation Module}
\label{sec:dataset}

We introduce a new design to generate a paired dataset for marine snow removal in underwater videos. This dataset enables supervised learning by providing aligned image pairs, where one frame is affected by marine snow, and the other is used as a clean reference. This process can be applied for footage made in different conditions, generating customised datasets that suit the needs of particular applications. Our dataset generation process is as follows:

First, we manually select underwater video sequences with minimal marine snow from the data available to us, which serve as our clean reference frames (ground truth). These sequences contain minimal visual noise, making sure they represent the true scene structure.

Then, we identify sequences that contain visible marine snow but do not have moving objects in the background. However, such sequences were not readily available in our dataset. To overcome this, we extract static background regions from snow-affected frames by selecting fixed-size patches of $550 \times 600$ pixels. These patches are chosen to ensure they contain snow particles while avoiding any foreground objects. To isolate the snow, we compute a median frame across each snow sequence and subtract it from each frame, producing a set of dynamic snow masks that capture the movement of marine snow particles. Given a sequence of $ N $ frames, denoted as $ I_t(x, y) $ where $ t \in \{1, 2, ..., N\} $, the median frame $ I_{\text{med}}(x, y) $ is computed as:
\begin{equation}
I_{\text{med}}(x, y) = \text{median} \left( I_1(x, y), I_2(x, y), \dots, I_N(x, y) \right).
\end{equation}

The snow mask sequence $ M_t(x, y) $ is then obtained by subtracting the median frame from each individual frame:
\begin{equation}
M_t(x, y) = I_t(x, y) - I_{\text{med}}(x, y).
\end{equation}

This operation removes the static background, leaving only the dynamic sequence with snow particles. Then, $ M_t(x, y) $ is overlaid onto the clean reference frames $ I_{\text{GT}}^{t'} $ to generate snow-affected frames $ I_{\text{snow}}^t(x, y) $. Since the extracted snow mask sequences may be shorter than the reference sequences, we apply a sliding window approach. The temporal alignment of the mask with the reference sequence is randomized, introducing natural variations in motion patterns. Additionally, the spatial position of the snow mask is randomized across the reference frame, ensuring variability in how marine snow appears in different locations: 
\begin{equation}
I^{\text{snow}}_t(x, y) = I^{\text{GT}}_{t'}(x + \Delta_x, y + \Delta_y) + M_t(x, y),
\end{equation}
where   $ \Delta_x, \Delta_y $ are spatial offsets used to randomize the overlay position. The temporal index $ t' $ is randomly selected within a sliding window, ensuring variation.

\begin{figure}
    \centering
    \includegraphics[width=\linewidth]{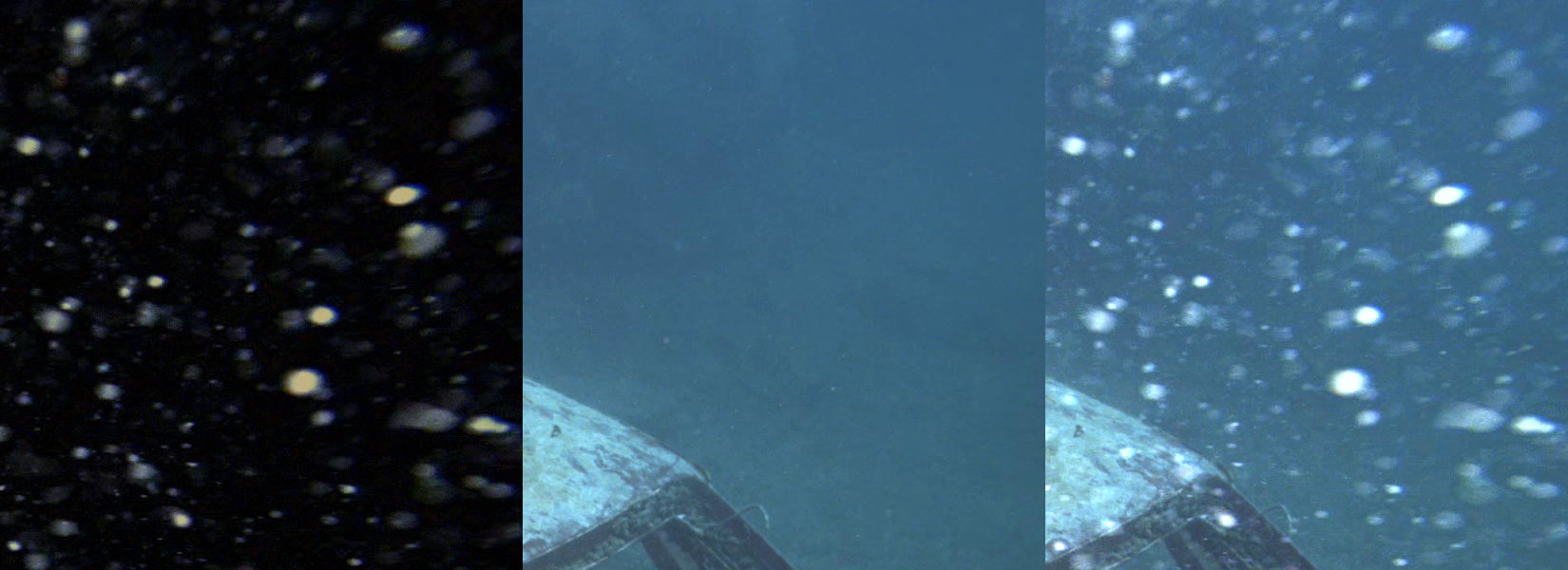}\vspace{2pt}
    \includegraphics[width=\linewidth]{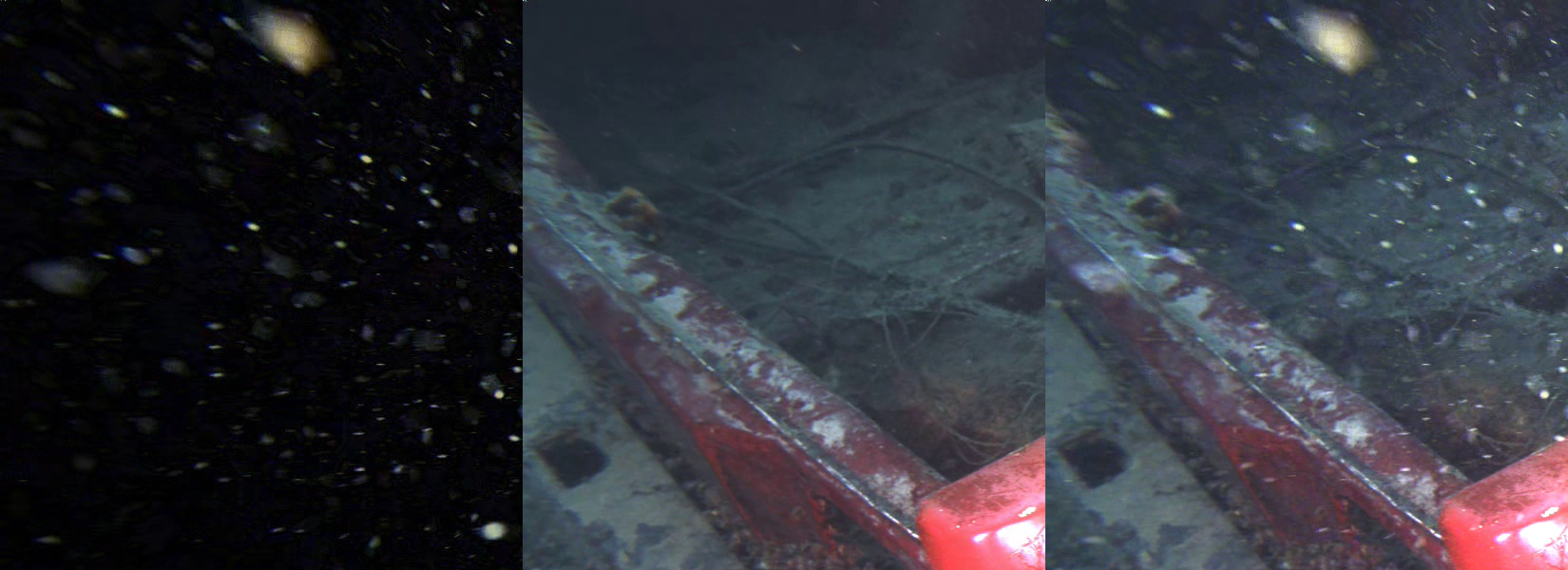}\vspace{2pt}
     \includegraphics[width=\linewidth]{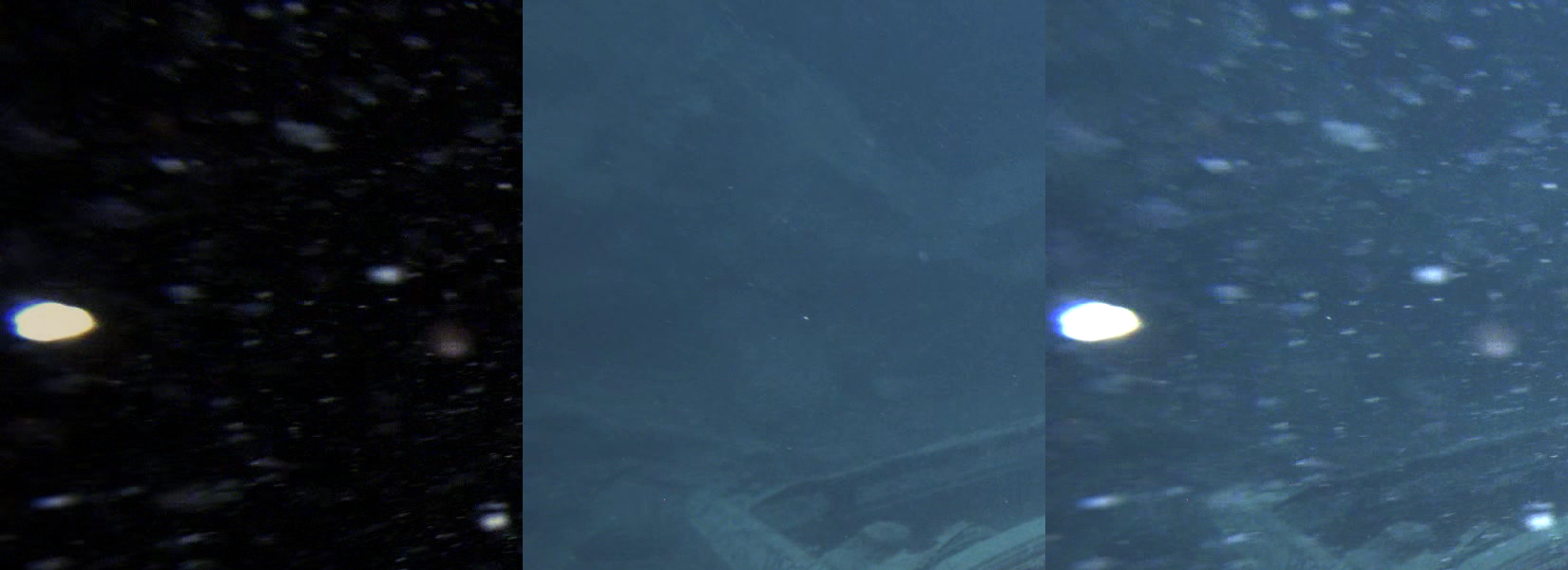}
    \caption{First column: extracted snow mask; second column: ground truth image; last column: resulting snow-affected image with overlaid snow mask. }
    \label{fig:dataset}
\end{figure}

As a result of this process, we generate a dataset of total 83,000+ paired frames, where each image pair consists of a snow-affected frame and a corresponding clean frame. The generated dataset contains 300 training sequences having a resolution of $550 \times 600$ pixels and covering a wide range of marine snow conditions and 10 testing sequences for evaluation. Fig. \ref{fig:dataset} shows the example frames from the dataset.

\begin{figure}
    \centering
    \scalebox{-1}[1]{\includegraphics[width=\linewidth]{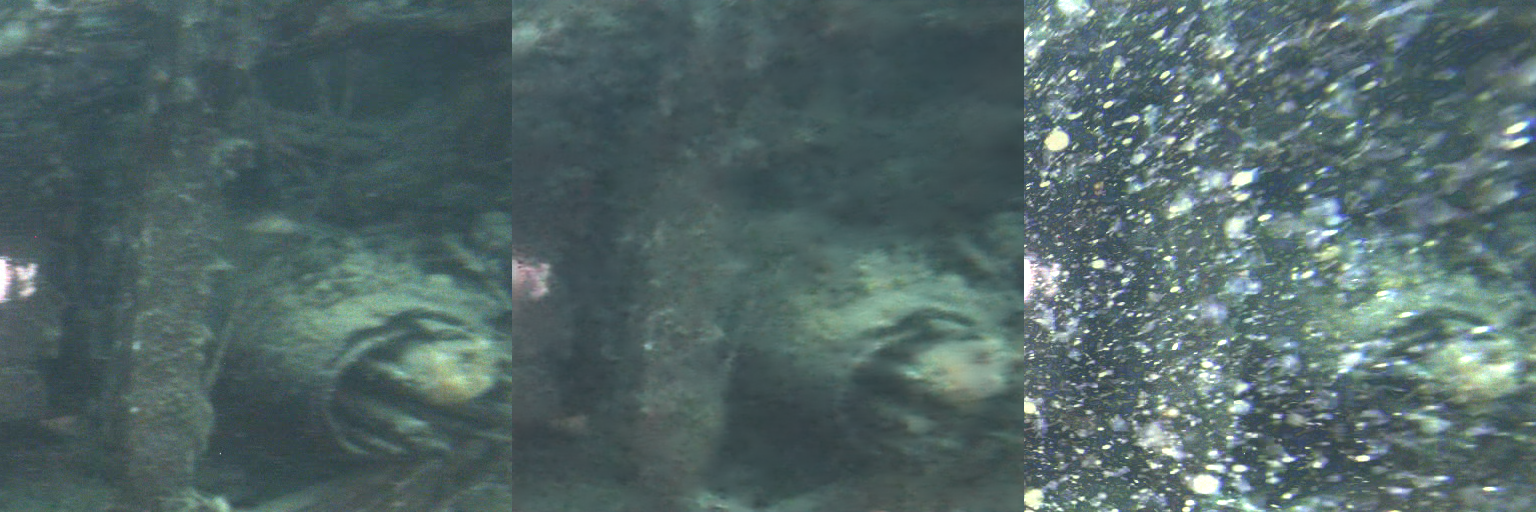}}\vspace{2pt}
     \scalebox{-1}[1]{\includegraphics[width=\linewidth]{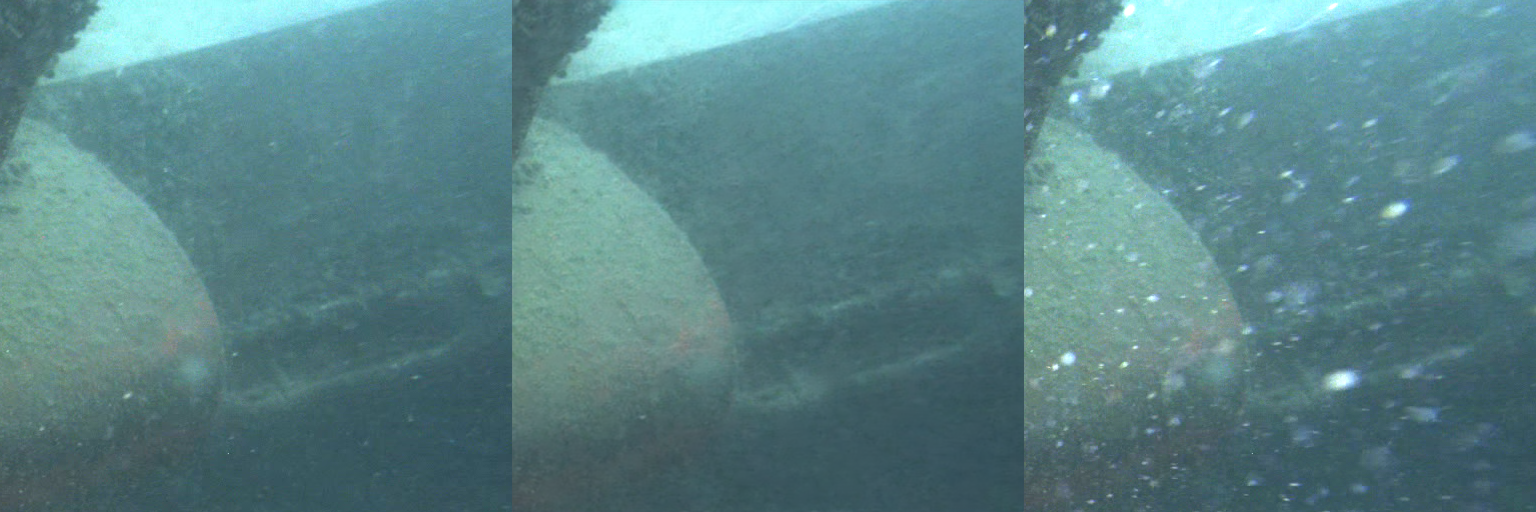}}\vspace{2pt}
    \scalebox{-1}[1]{\includegraphics[width=\linewidth]{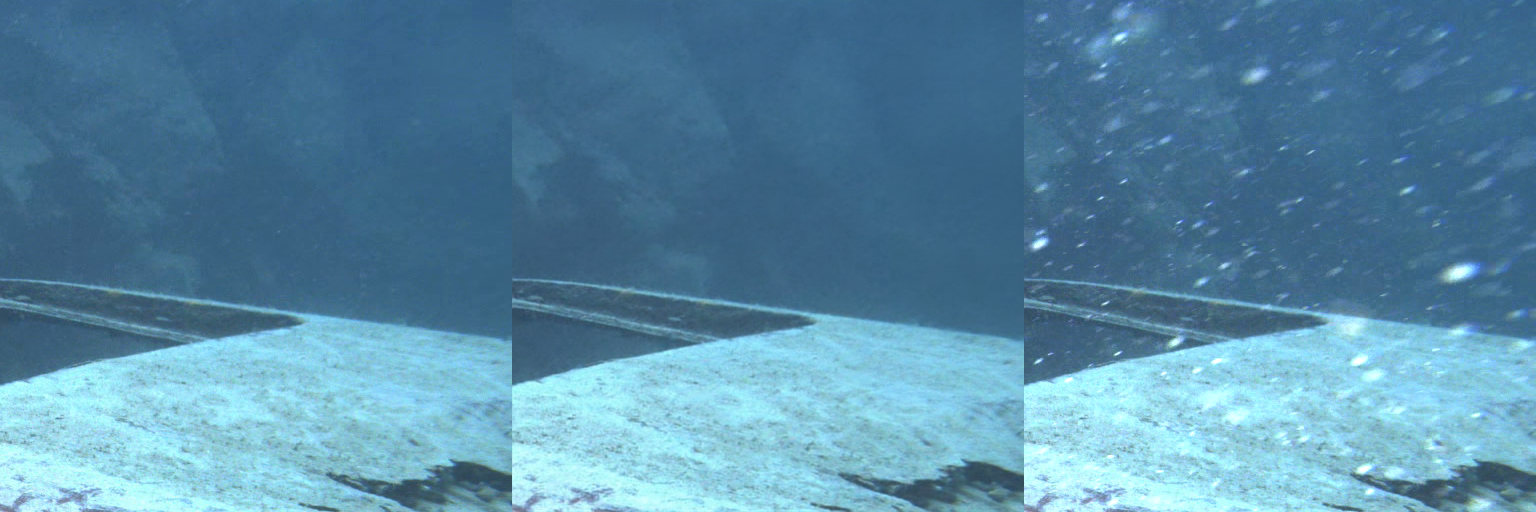}}
    \caption{Subjective results on test set. First column: Raw input image with real marine snow; second column: result; last column: ground truth image. }
    \label{fig:subj_eval}
\end{figure}

\subsection{Video Enhancement Module}

We employ BVI-Mamba \cite{lin2024bvi}, an architecture that incorporates feature-level frame alignment and integrates State Space Models (SSMs) with the 2D Selective Scan (SS2D) module \cite{gu2023mamba}. This design efficiently models long-range dependencies while maintaining a computationally efficient framework. The model captures temporal correlations across frames with linear complexity, making it effective for handling long sequences while preserving fine spatial details.  

\section{Experiments and Results}
\label{sec:results}
\subsection{Datasets and Training Setup}

Underwater sequences were provided by Beam\footnote{https://beam.global/}, a company that uses underwater robots to service offshore wind farms. These sequences are in RGB format with resolution $2048\times1080$, captured using Beam’s mapping system in both commercial and controlled testing environments. Each sequence includes inspections of submerged structures, such as jacket members, anodes, and a fuselage, with durations ranging from 8.5 minutes to 2 hours 39 mins. For our experiments, shorter 1-2-minute clips are extracted from the full sequences.

Eight sequences for task-specific performance evaluation, each containing between 1,241 and 2,841 frames, were kept separate from the main dataset for further task-based evaluation.

For training, we used  $512\times512$ patch size, input sequence length of 5 frames, learning rate of 0.0001 and $l_1$ loss. The training results on the test set are shown in Fig. \ref{fig:subj_eval}.

\subsection{Performance Metrics}

Evaluating enhancement performance with a generated dataset like ours (even if that is derived from real-world data) is challenging. Since our paired data is derived from frame blending, the "ground truth" is only the cleanest available data, not a truly clean underwater scene. This makes it difficult to assess whether enhancement restores true lost details or introduces artifacts. 
Also, standard objective image quality metrics like PSNR and SSIM that rely on pixel-wise differences may not provide meaningful insights into real-world performance. Pixel-level performance does not necessarily correlate with the performance of downstream tasks, like SLAM, which depend on preserving key visual features rather than exact pixel-level reconstruction.

Therefore, a better approach is to assess the impact of video enhancement on SLAM performance. Since SLAM relies on feature tracking, matching, spatial and temporal consistency, improvements in these aspects provide a more relevant measure of the quality of the enhancement. If an enhancement method improves SLAM robustness in challenging conditions under water, it is likely more useful than one that optimizes PSNR, SSIM or other objective pixel-level metrics on generated test data.

For performance evaluation we chose three task-relevant metrics: the number of keypoints per frame in SLAM, the number of frame-to-frame feature matches in SLAM and the number of points in the reconstructed 3D-models (dense mapping).

\subsection{Results}

We compared the performance of Beam's proprietary mapping system, which supports real-time, high-resolution, and globally consistent mapping; the system includes a feature extraction and matching module used as part of the mapping pipeline. The SLAM and dense mapping systems are run on the original sequences without enhancement, the sequences enhanced with the Bayesian enhancement model \cite{huang2025bayesian} and our method. We chose 8 real video sequences to evaluate the performance. The sequences starting with 2022\_ are affected by heavy marine snow, whereas the sequences starting with 2023\_ have low levels of snow.

\begin{figure}
    \centering
    \includegraphics[width=\linewidth]{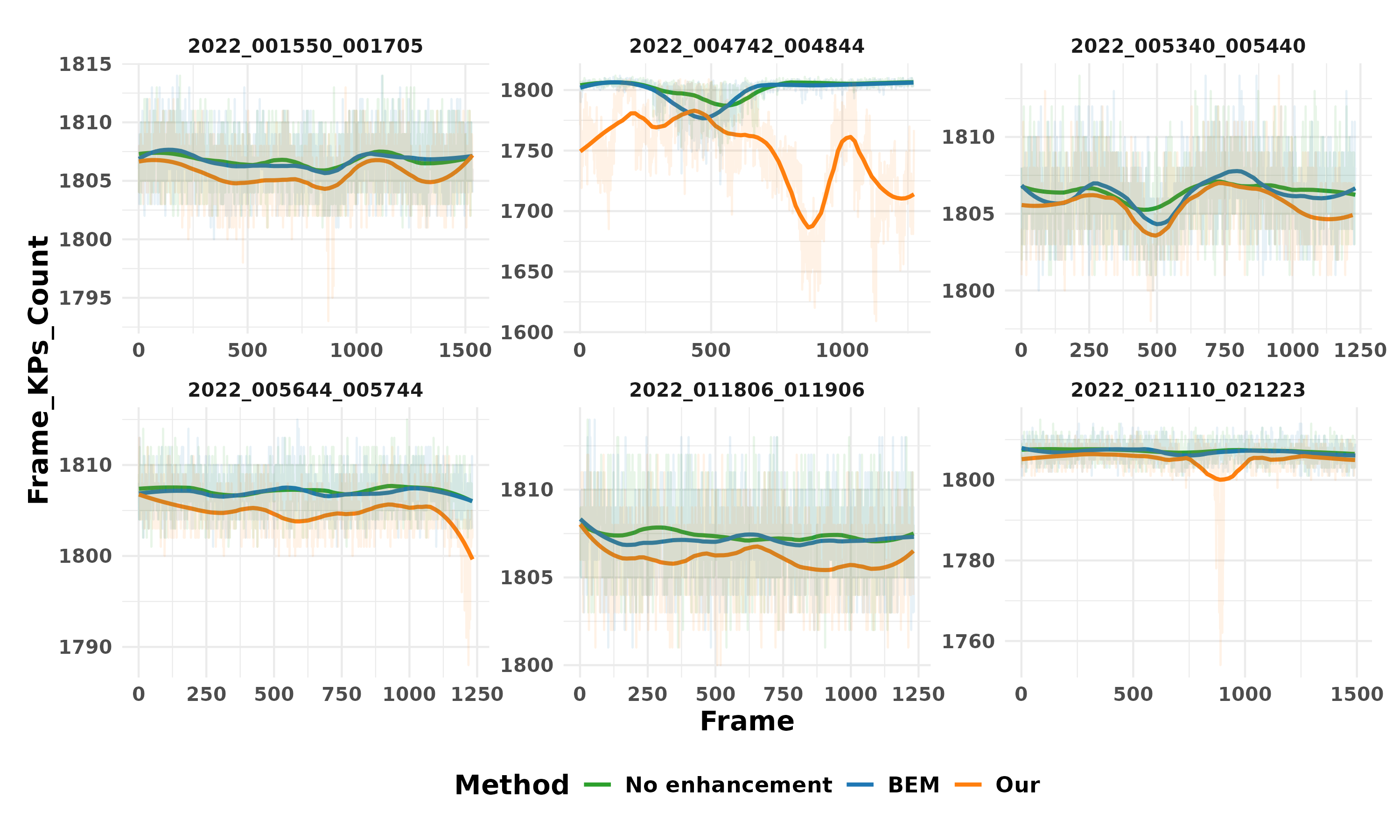} 
    \vspace{-20pt}
    \caption{Number of keypoints per frame (lower is better, indicating fewer snow particles). Smoothing was applied to the data for clarity.}
    \label{fig:num_keypts}
\vspace{10pt}
    \centering
    \includegraphics[width=\linewidth]{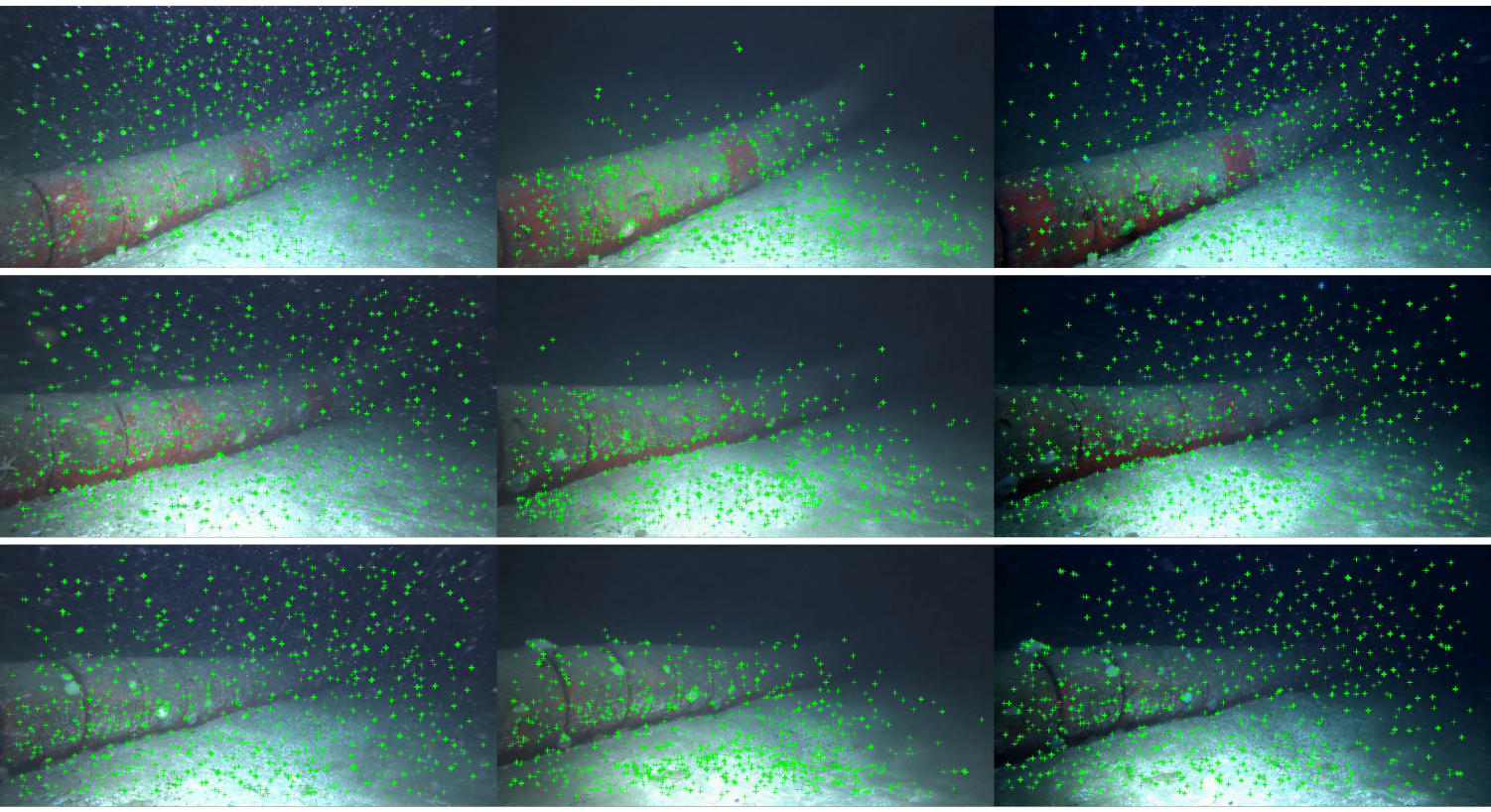}
    \vspace{-20pt}
    
    \caption{Example frames and feature detection (feature points shown in green) for sequence 2022$\_$004742$\_$004844. Feature detection is performed on left column: no enhancement; central column: our method; right column: BEM.}
    \label{fig:features}
\vspace{10pt}
    \centering
    \includegraphics[width=\linewidth]{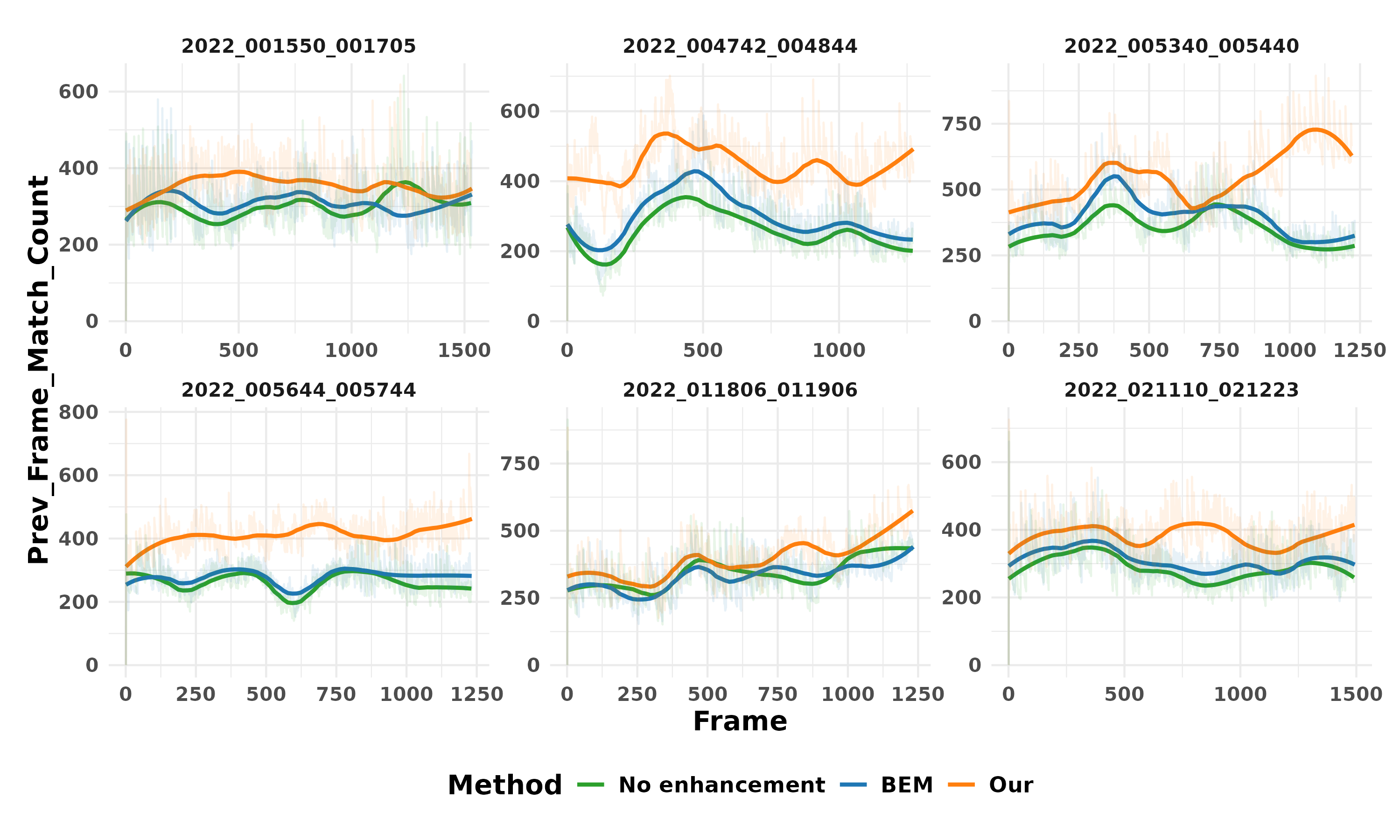}  \vspace{-20pt}
    \caption{Number of frame-to-frame feature matches (higher is better, indicating more useful features for SLAM). Smoothing was applied to the data for clarity.}
    \label{fig:num_f2f_matches}
\end{figure}

Fig. \ref{fig:num_keypts} shows the number of keypoints per frame, where high numbers indicate high levels of snow. Overall, 
our results show a lower number of keypoints compared to sequences without enhancement and those enhanced with BEM. That suggests that our method may suppress noise caused by snow particles. The example of the feature point detection is shown in Fig. \ref{fig:features}. Note how for our method, features are focused mainly on the seabed and cable regions rather than snow particles.

Our method also consistently achieves a higher number of frame-to-frame matches, as shown in Fig. \ref{fig:num_f2f_matches}, that means the preserved or enhanced features are more consistent across frames, improving feature tracking stability in SLAM. This suggests that although fewer keypoints are detected, they are of higher quality and better suited for matching.

\begin{figure}
    \centering
    \includegraphics[width=\linewidth]{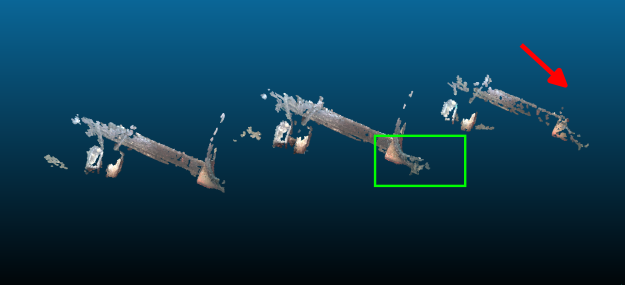}\vspace{2pt}
    \includegraphics[width=\linewidth]{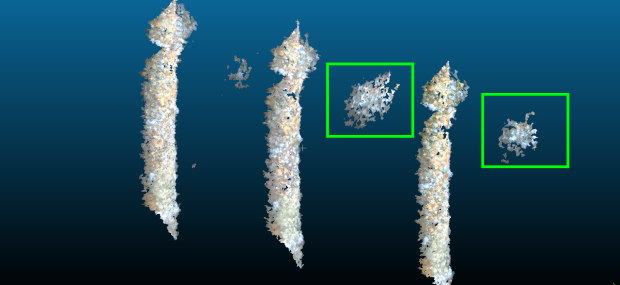}
    
    \caption{3D models (point cloud) reconstructed for sequence 2023$\_$000015$\_$000135 (top) and sequence 2022\_005644\_005744. Methods order (left to right): no enhancement, our method, BEM. The red arrow highlights key differences (missing parts or sparsity) of the enhanced video with respect to the non-enhanced baseline. The green box highlights improved reconstruction with respect to the baseline}
    \label{fig:3d_example}
\end{figure}

Table \ref{tab:average_points_3d} demonstrates the result for average number of points in dense mapping. Our approach results in a higher average number which demonstrates improved scene reconstruction quality. This implies that the enhancement reduces noise and artifacts that would otherwise disrupt dense geometry estimation, leading to more reliable 3D models. Figure \ref{fig:3d_example} shows the example of the reconstructed 3D model (point cloud) for the sequence 2022$\_$004742$\_$004844. Our method shows details not appearing on the reconstructed model obtained from a non-enhanced sequence and sequence processed with BEM.

\begin{table}
    \centering
    \renewcommand{\arraystretch}{1.2}
    \caption{Average number of points per method in 3D-reconstruction}
    \label{tab:average_points_3d}
    \begin{tabular}{|c|c|}
        \hline
        \textbf{Method} & \textbf{Average Num of Points } \\ \hline
        No Enhancement & 893,076 \\ \hline
        BEM & 802,084.6 \\ \hline
        Our & 1,142,521.5 \\ \hline
    \end{tabular}
\end{table}

\section{Conclusion}
\label{sec:conclusion}
This paper introduced a novel dataset generation approach and a video enhancement method for marine snow removal in underwater videos. The framework generates a paired dataset from real snow artefacts, providing correct characteristics of distortion for the enhancement network. Our results show that the proposed method improves SLAM performance by enhancing key visual features while preserving spatial and temporal consistency. This work provides a practical solution for underwater vision tasks affected by marine snow.

\bibliographystyle{IEEEtran}
\bibliography{references} 
\FloatBarrier
\end{document}